\documentclass[12pt]{article}
\usepackage{geometry}
\usepackage{setspace}
\usepackage{fancyhdr}
\usepackage{amsfonts}
\usepackage{amsmath}
\usepackage{verbatim}
\usepackage{listings}
\usepackage{lscape}
\usepackage{graphics}
\usepackage{tabularx}
\usepackage{float}
\usepackage{breqn}
\usepackage{wasysym}
\usepackage{fix-cm}
\usepackage{bbm}
\usepackage{color}
\usepackage{amsthm}
\usepackage{enumitem}
\usepackage{authblk}

\usepackage{appendix}

\usepackage{graphicx}

\usepackage{multirow}
\usepackage{parskip}
\usepackage{amsmath, amsthm, amssymb}
\usepackage{dsfont}
\usepackage{bm}
\usepackage{parskip}

\usepackage{framed,enumitem}
\usepackage{tikz}
\usetikzlibrary{positioning}
\usepackage{accents}
\usepackage{comment}
\usepackage[ruled,vlined]{algorithm2e}
\usepackage{breqn}

\usepackage[round]{natbib}

\newtheorem{myprop}{Proposition}
\theoremstyle{definition}
\newtheorem{mydef}{Definition}
\newenvironment{example}[1][Example]{\begin{trivlist}
\item[\hskip \labelsep {\bfseries #1}]}{\end{trivlist}}

\makeatletter
\newtheorem*{rep@theorem}{\rep@title}
\newcommand{\newreptheorem}[2]{%
\newenvironment{rep#1}[1]{%
 \def\rep@title{#2 \ref{##1}}%
 \begin{rep@theorem}}%
 {\end{rep@theorem}}}
\makeatother

\newreptheorem{theorem}{Theorem}
\newreptheorem{lemma}{Lemma}
\newreptheorem{mycor}{Corollary}
\newreptheorem{myprop}{Proposition}

\providecommand{\keywords}[1]
{
  \small	
  \textbf{\textit{Keywords---}} #1
}

\title{
A note on incorrect inferences in non-binary qualitative probabilistic networks
}
\author{Jack Storror Carter}
\affil{Dipartimento di Matematica, Universit\`a degli Studi di Genova, Genova, Italy}
\affil{Correspondence to jackcarter23@aol.com}
\date{}

\begin{document}

\maketitle

\begin{abstract}
    Qualitative probabilistic networks (QPNs) combine the conditional independence assumptions of Bayesian networks with the qualitative properties of positive and negative dependence.  They formalise various intuitive properties of positive dependence to allow inferences over a large network of variables.
    However, we will demonstrate in this paper that, due to an incorrect symmetry property, many inferences obtained in non-binary QPNs are not mathematically true.  We will provide examples of such incorrect inferences and briefly discuss possible resolutions.
\end{abstract}


\section{Introduction}

Introduced by \cite{Wellman1990}, qualitative probabilistic networks (QPNs) generated a fairly substantial literature with a number of papers published in journals and conference proceedings over the last 30 years.  They combine qualitative assumptions with the conditional independence assumptions of Bayesian networks.  A Bayesian network often makes strong distributional assumptions about the associated variables with variables assumed to come from a particular family of distributions.  \cite{Wellman1990} argued that such assumptions are `inappropriately precise for many applications' and can lead to them being `applicable in only narrow domains.'  In contrast, weaker qualitative assumptions can be less restrictive and still provide a framework where useful properties can be deduced, while also being more robust when elicited from expert judgements \citep{VanderGaag2010}.

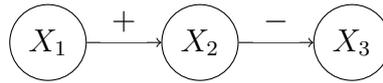
\begin{figure}
    \centering
    \begin{tikzpicture}[node distance={20mm},main/.style = {draw, circle}] 
    \node[main] (1) {$X_1$}; 
    \node[main] (2) [right of=1] {$X_2$};
    \node[main] (3) [right of=2] {$X_3$};
    \draw[->] (1) -- (2) node [midway, above, fill=white] {$+$};
    \draw[->] (2) -- (3) node [midway, above, fill=white] {$-$};
    \end{tikzpicture} 
    \caption{An example QPN.}
    \label{fig:graph}
\end{figure}

The primary qualitative property specified by a QPN is positive and negative dependence, also called qualitative influences in the literature.  These qualitative influences are represented visually by a QPN via a signed directed acyclic graph (DAG) where the vertices of the graph represent variables.  The lack of an edge specifies some (conditional) independence relationship between the variables, while the directed edges are each labeled with a $+$, $-$ or $?$ sign.  A $+$ edge represents a positive influence between the corresponding variables, a $-$ edge represents a negative influence and a $?$ edge represents an unspecified influence (a dependence which is not necessarily either positive or negative).  For example, the graph in Figure \ref{fig:graph} represents a QPN for the variables $(X_1,X_2,X_3)$.  This QPN specifies that $X_1$ and $X_3$ are conditionally independent given $X_2$, $X_1$ has a positive influence on $X_2$, and $X_2$ has a negative influence on $X_3$.

QPNs allow inferences to be made over large networks of variables - inferences that would not be intuitively obvious without the mathematical framework.  For example, one use of a QPN is when some variables are under a decision makers control and another variable is an objective \citep{Wellman1990,Renooij1998}.  The QPN can be used to determine if the decision variables have a positive or negative influence on the objective.

Most works in the QPN literature assume that qualitative influences satisfy a symmetry property - if $X_1$ positively influences $X_2$ then a new QPN can be constructed in which $X_2$ positively influences $X_1$.  This symmetry is vital for many of the inferences made by QPNs \citep{Druzdzel1993}.  However, we will demonstrate in this paper that this symmetry property does not hold for non-binary variables and therefore many inferences are not mathematically true.

The remainder of the paper is organised as follows.  In Section \ref{sec:QPNs} we formally define a QPN and give a brief literature review.  In Section \ref{sec:PD} we discuss positive dependence and demonstrate how the current definition of positive dependence is not symmetric.  In Section \ref{sec:reversals} we show how this mistake affects inferences in QPNs and in Section \ref{sec:ex} we give an applied example of an incorrect inference.  We finish in Section \ref{sec:disc} with a discussion.

\section{QPNs}\label{sec:QPNs}

A QPN is specified by a signed DAG $\mathcal{G} = (V, E)$ where the vertex set $V$ corresponds to random variables $X = (X_{v})_{v \in V}$ and $E \subset \{ (i,j,\delta) : i,j \in V, \delta \in \{ +, -, ? \} \}$ is the signed edge set.  Under the QPN it is assumed that the variables $X$ follow certain conditional independence relationships specified by the graph $\mathcal{G}$ via the global Markov property.  See, for example, \cite{Maathuis2018}, \cite{darwiche2009} and \cite{koller2009} for comprehensive overviews of graphs, graphical models and conditional independence.

The signed edges in a QPN also stipulate certain positive and negative dependence relationships between the variables.  Within QPNs, positive dependence is defined via first-order stochastic dominance (FSD).

\begin{mydef}\label{def:FSD}
Let $F$ and $F'$ be two cumulative density functions with the same support $A \subseteq \mathbb{R}$.  We say that $F$ FSD $F'$ if $F(x) \leq F'(x)$ for all $x \in A$.
\end{mydef}

Informally, if $F$ FSD $F'$ then larger values are more likely under $F$ than under $F'$.  FSD gives a partial ordering of cumulative density functions (cdfs).  For more information on FSD and examples see \cite{levy2015}.  This is used to define positive dependence between two variables $X_i$ and $X_j$ through conditioning on larger values of $X_i$ leading to FSD of the cdf of $X_j$ - in other words, observing a large value of $X_i$ makes larger values of $X_j$ more likely.  Due to the graphical structure of the QPN, this must hold when conditioning on any value of the \textit{parents} of $X_j$.  A parent of the vertex $j$ in graph $\mathcal{G}$ is any vertex $k$ with an edge $(k,j,\delta) \in E$ directed towards $j$.

\begin{mydef}\label{def:PosInf}
Let $F_{X_j \mid x_i, x_K}$ be the cdf of $X_j$ given $X_i=x_i$ and $X_K=x_K$ where $K$ is the set of parents of $j$ except for $i$.  We say that:
\begin{itemize}
\item $X_i$ positively influences $X_j$ in the graph $\mathcal{G}$ and we write $S^{+}(i,j,\mathcal{G})$ if $F_{X_j \mid x_i, x_K}$ FSD $F_{X_j \mid x'_i, x_K}$ for all $x_i > x'_i$ and all $x_K$.

\item $X_i$ negatively influences $X_j$ in the graph $\mathcal{G}$ and we write $S^{-}(i,j,\mathcal{G})$ if $F_{X_j \mid x_i, x_K}$ FSD $F_{X_j \mid x'_i, x_K}$ for all $x_i < x'_i$ and all $x_K$.
\end{itemize}
\end{mydef}

Under a QPN $\mathcal{G}$ it is assumed that for any $(i,j,+) \in E$ we have $S^{+}(i,j,\mathcal{G})$, and for any $(i,j,-) \in E$ we have $S^{-}(i,j,\mathcal{G})$.  If these, as well as the conditional independence relationships, hold for $X$, then we say that $X$ satisfies the QPN $\mathcal{G}$.  In the literature, $\mathcal{G}$ is often omitted from $S^{+}(i,j,\mathcal{G})$ when the QPN is known from context.

There is a further level to QPNs of additive and product synergies which can be represented graphically by a hyperedge.  Synergies are used to `describe the qualitative interaction among influences' \citep{Wellman1990}, and can be useful, for example, in the case of competing causes of an event \citep{Wellman1993, Druzdzel1993a}.  However, for the purposes of this paper synergies are not required and so we will focus solely on the positive and negative influences.

The main use of a QPN is to determine positive or negative influences between variables that are not directly connected in the graph.  Because the variables are not directly connected they are conditionally independent given the remaining variables.  However, the variables may still be marginally dependent and a QPN can be used to determine if the variables are positively or negatively dependent in their marginal distribution.  This may be to find the influence of a decision variable on an objective variable, or to propagate the effect of an observation of a variable through the graph.

After being introduced by \cite{Wellman1990}, there has been a significant number of publications working to improve the methodology and expand the framework of QPNs.  Notable research includes improved inference algorithms \citep{Wellman1990a,Druzdzel1993,Renooij1998,Renooij2002,VanKouwen2009}, ambiguity resolution \citep{Parsons1995,Renooij2000,Renooij2000a}, influence strengths \citep{Renooij1999,Renooij2003,Renooij2008,Yue2010,Yue2015}, context dependent signs \citep{Renooij2000b,Renooij2002b,Bolt2005} and semi qualitative probabilistic networks \citep{Renooij2002a,DeCampos2005,DeCampos2009,Campos2013}.
Most recently, QPNs have been used to help construct probabilistic relational models \citep{VanDerGaag2018} and were studied by \cite{Lv2019} who aimed to combine the information of multiple QPNs through sign fusion.

\section{Positive dependence}\label{sec:PD}

While positive dependence in QPNs requires conditioning on the parent variables, as in Definition \ref{def:PosInf}, for simplicity we will ignore this conditioning in this section.  However, all discussion and results in this section remain valid with the conditioning included.

A key decision in the development of QPNs was the choice of Definition \ref{def:PosInf} to define a positive dependence relationship between two variables in the graph.  However, there are many other proposed definitions of positive dependence - see, for example, \cite{Lehmann1966} and \cite{Colangelo2005} for overviews.  The simplest says two random variables are positively dependent if they have positive correlation.  This is generally a weak form of positive dependence and other more strong definitions have been proposed.  For example, the random variables $X$ are called \textit{associated} if $\mathrm{cov}(f(X),g(X)) \geq 0$ for any two non-decreasing functions $f$ and $g$ for which $\mathbb{E}|f(X)|$, $\mathbb{E}|g(X)|$ and $\mathbb{E}|f(X)g(X)|$ all exist \citep{Esary1967}.  An even stronger definition is total positivity of order 2 (TP$_2$).  The variables $(X,Y)$ with joint density function $p$ are said to be TP$_2$ if $p(x,y')p(x',y) \leq p(x,y) p(x',y')$ for all $x<x'$, $y<y'$.

Each of these positive dependence properties are quite clearly symmetric.  
This symmetry is even evident in the language we use to describe them - one would generally say that $X$ and $Y$ are positively correlated.  
However, the directionality inherent to the graphical model of a QPN seems to lend itself more to a definition of positive definiteness which is not so obviously symmetric.  Indeed, this is the approach taken in QPNs with the use of positive influences.  This definition of positive dependence is in fact not symmetric (see example below) and this is reflected in the language where we say that X positively influences Y.

A second form of positive dependence considered by \cite{Wellman1990} is the monotone likelihood ratio property (MLRP).

\begin{mydef}
Let $p_{X|Y}( \cdot \mid y)$ denote the probability density function of $X$ given $Y=y$.  $X$ and $Y$ satisfy the MLRP if for any $x \geq x'$ and $y \geq y'$,
$$ \frac{p_{X|Y}( x \mid y )}{p_{X|Y}( x \mid y' )} \geq \frac{p_{X|Y}( x' \mid y )}{p_{X|Y}( x' \mid y' )} $$
\end{mydef}

The MLRP is in fact equivalent to $X$ and $Y$ being TP$_2$, as long as $Y$ has non-zero density across its whole support.  This can easily be seen by replacing the conditional densities by $p_{X|Y}(x \mid y) = \frac{p_{X,Y}(x,y)}{p_{Y}(y)}$ where $p_{X,Y}$ denotes the joint density of $(X,Y)$ and $p_{Y}$ denotes the marginal density of $Y$.  This demonstrates that the MLRP is also a symmetric property.

The reason for the name monotone likelihood ratio property comes from considering the distribution of $Y \mid X=x$ as a posterior.  Then the marginal distribution of $Y$ can be interpreted as a prior and the distribution of $X \mid Y=y$ as a likelihood.  On the other hand, $X$ positively influencing $Y$ can be seen as a property on the posterior distributions of $Y \mid X=x$.  This prior and posterior interpretation leads to the following relationship between positive influences and the MLRP, which is an adaptation of a result in \cite{Milgrom1981}.

\begin{myprop}\label{prop:Milgrom}
$X$ and $Y$ satisfy the MLRP if and only if $X$ positively influences $Y$ for any choice of prior distribution on $Y$.
\end{myprop}

The important part of Proposition \ref{prop:Milgrom} is that the positive influence must hold for \textit{any} choice of prior distribution on $Y$.  Hence the MLRP always implies that $X$ positively influences $Y$ (and also that $Y$ positively influences $X$), but the inverse is not true.  However, due to a misinterpretation of Proposition \ref{prop:Milgrom}, \cite{Wellman1990} uses the definitions of positive influence and the MLRP interchangeably.  Later, \cite{Druzdzel1993} incorrectly stated positive influence as a symmetric property, using the false equivalence with the MLRP to prove this.  This incorrect symmetry property was subsequently widely adopted in the QPN literature.  

We now give an example demonstrating both the lack of equivalence between positive influence and the MLRP and the lack of symmetry of positive influences (see \cite{Lehmann1966} Section 5 for further examples of the lack of symmetry).

\begin{example}
Let $X$ and $Y$ be two trivariate random varibles taking values in $\{ 1,2,3 \}$.  Suppose the joint probabilities of $(X,Y)$ are as in Table \ref{tab:ex1}.

\begin{table}[h]
\centering
\begin{tabular}{clclll}
                   &                        & \multicolumn{3}{c}{Y}                     &       \\
                   & \multicolumn{1}{l|}{}  & 1     & 2    & \multicolumn{1}{l|}{3}     &       \\ \cline{2-6} 
\multirow{3}{*}{X} & \multicolumn{1}{l|}{1} & 0.2   & 0.05 & \multicolumn{1}{l|}{0.075} & 0.325 \\
                   & \multicolumn{1}{l|}{2} & 0.15  & 0.15 & \multicolumn{1}{l|}{0.1}   & 0.4   \\
                   & \multicolumn{1}{l|}{3} & 0.075 & 0.1  & \multicolumn{1}{l|}{0.1}   & 0.275 \\ \cline{2-6} 
                   & \multicolumn{1}{l|}{}  & 0.425 & 0.3  & \multicolumn{1}{l|}{0.275} &      
\end{tabular}
\caption{Joint probabilities for $(X,Y)$.}
\label{tab:ex1}
\end{table}

By considering the relevant conditional probabilities, it is easy to verify that in this example $X$ positively influences $Y$ but $Y$ does not positively influence $X$.  This demonstrates that positive influence is not a symmetric property.

Furthermore, this distribution of $(X,Y)$ does not satisfy the MLRP because, for example,
$$ \frac{ \mathbb{P}( X = 3 | Y = 3 ) }{ \mathbb{P}( X = 3 | Y = 2 ) } \approx 1.09 < 1.64 \approx \frac{ \mathbb{P}( X = 1 | Y = 3 ) }{ \mathbb{P}( X = 1 | Y = 2 ) } $$
Hence $X$ positively influencing $Y$ does not imply that $(X,Y)$ satisfy the MLRP.

\end{example}

One caveat to this discussion is for when $X$ and $Y$ are both binary variables.  In this case it is easy to show that the MLRP and positive influence are equivalent.  In fact, for binary variables all definitions of positive dependence are equivalent.

\section{Inference in QPNs}\label{sec:reversals}

Inference in QPNs consists of determining positive and negative influences between variables that are not directly linked in the QPN.  That is, if $X$ satisfies the QPN $\mathcal{G}$, then what can we say about the marginal distribution of any pair of variables in $X$?  In the literature there have been two main approaches for inference algorithms.  The first, often referred to as query processing, aims to find the direction of influence of some conditioning variables on a target variable.  A method, proposed by \cite{Wellman1990} and developed further in \cite{Wellman1990a}, consists of implementing a sequence of \textit{reductions} and \textit{reversals} to the QPN until only the relevant variables remain and all edges between the conditioning variables and the target variable are directed towards the target variable.

A reduction removes a vertex $i$ and finds a new QPN with vertex set $V' = V \setminus \{ i \}$ which the remaining variables $(X_v)_{v \in V \setminus \{ i \} }$ satisfy.  It was shown that this could be done when $i$ has at most one parent in $\mathcal{G}$ with the signs of new edges determined by two operators (\cite{Wellman1990}, Theorem 4.3).

A reversal aims to find a new QPN $\mathcal{G}'$ which $X$ satisfies but with an edge $(i,j,\delta) \in E$ reversed such that the new edge set contains $(j,i,\delta') \in E'$.  It was claimed that the sign of the reversed edge can remain unchanged (i.e. $\delta = \delta'$), but that the signs of certain other edges might need to be changed to $?$ (\cite{Wellman1990}, Theorem 4.4).

The second approach to inference in QPNs is often referred to as message passing and aims to find the effect of an observation of a single variable on the remaining variables.  First the observation is determined to be either large or small.  Then a message passing algorithm propagates the influence of this observation through the network, beginning at the observed vertex and passing along the edges of the graph.  In this way each vertex is assigned either a $+$, $-$ or $?$ which shows if the observation makes larger values more likely, smaller values more likely or has an undetermined influence.  A message passing algorithm was first developed by \cite{Druzdzel1993} with improved algorithms proposed by \cite{Renooij1998} and \cite{VanKouwen2009} and an algorithm for observations of multiple variables by \cite{Renooij2002}.

\begin{figure}[h]
    \centering
    \begin{tikzpicture}[node distance={20mm},main/.style = {draw, circle}] 
    \node[main] (1) {$X$}; 
    \node[main] (2) [right of=1] {$Y$};
    \draw[->] (1) -- (2) node [midway, above, fill=white] {+};
    \end{tikzpicture} 
    \caption{A very simple QPN in which $X$ positively influences $Y$.  However, as demonstrated in Section \ref{sec:PD}, we cannot conclude that $Y$ also positively influences $X$.}
    \label{fig:graph2}
\end{figure}
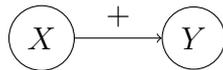

To demonstrate the potential error in these inference algorithms when applied to non-binary QPNs, we will give a very simple example.  Consider the QPN in Figure \ref{fig:graph2} which consists of two variables and a positive influence from $X$ to $Y$.  Suppose we wish to determine the influence of $Y$ on $X$.  To do this, a query processing algorithm will reverse the direction of the edge while maintaining the $+$ sign (following \cite{Wellman1990}, Theorem 4.4) and conclude that $Y$ has a positive influence on $X$.  

Now instead suppose that we observe a large value of $Y$ and wish to know the effect this has on $X$.  All message passing algorithms will pass a message along the edge from $Y$ to $X$, matching it's sign, and determine that this observation of $Y$ makes larger values of $X$ more likely.

These conclusions are correct when both variables are binary.  However, as demonstrated in the example of Section \ref{sec:PD}, these conclusions are not true for non-binary variables.  In particular, both edge reversal (for query processing algorithms) and message passing in the reverse direction to an edge (for message passing algorithms) rely on the property of symmetry of influences - without symmetry the conclusion is no longer valid.

To the best of the author's knowledge, all current inference algorithms rely on the ability to reverse an edge or pass a message in the opposite direction to an edge.  Hence all algorithms, when applied to non-binary QPNs, can suffer from this incorrect inference.  This includes the inference algorithms of \cite{Wellman1990}, \cite{Wellman1990a}, \cite{Druzdzel1993}, \cite{Renooij2002} and \cite{VanKouwen2009}.  The algorithm of \cite{Renooij1998} would be similarly affected, however in this paper it is specifically stated that all variables should be binary.  It can also effect inference algorithms for extensions of non-binary QPNs.  For example, the incorrect inference can occur for context dependent signs \citep{Renooij2002b} and semi qualitative probabilistic networks \citep{DeCampos2005,DeCampos2009}.  For other extensions which would be affected, such as \cite{Renooij2000b} for ambiguity resolution, \cite{Renooij2003} for influence strengths, and \cite{Renooij2002a} for semi qualitative probabilistic networks, it is unclear whether the methodology is intended only for binary variables.

Such incorrect inferences can also occur in larger networks and lead to further incorrect conclusions that are not directly related to the incorrect reversal or symmetry of an edge.  We demonstrate this in the following example.

\section{Example}\label{sec:ex}

\begin{figure}[h]  
    \centering
\begin{tabular}{cc}
	\includegraphics[scale=0.55]{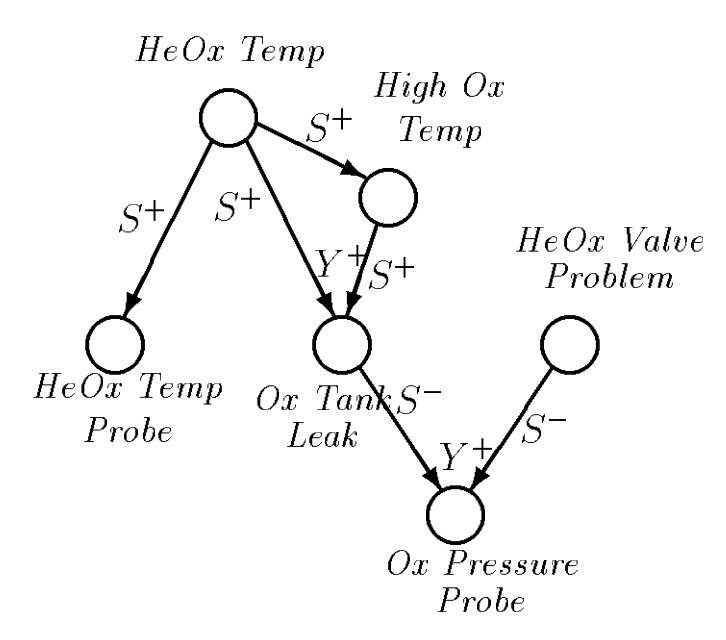} &
	\includegraphics[scale=0.55]{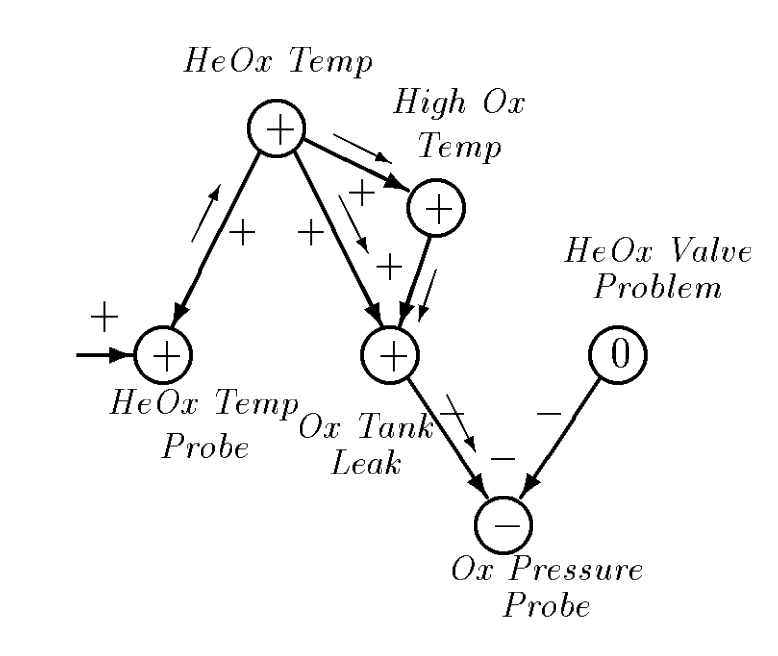}
\end{tabular}
    \caption{Left: A QPN for space shuttle example.  Right: Results of a message passing algorithm after observing a large value for \textit{HeOx Temp Probe}.  Both figures originally appeared in \cite{Druzdzel1993}.}
    \label{fig:graph3}
\end{figure}

We now investigate an example taken from \cite{Druzdzel1993}, in turn based on a belief network proposed in \cite{Horvitz1992}, which models the orbital maneuvering system propulsion engine of a space shuttle.  Of interest here is the temperature in the neighbourhood of two tanks (\textit{HeOx Temp}) which is measured by a probe (\textit{HeOx Temp Probe}).  A high temperature in the neighbourhood of the tanks can cause a high temperature in the oxidizer tank (\textit{High Ox Temp}), both of which can cause a leak in the oxidizer tank (\textit{Ox Tank Leak}).  A leak may lead to decreased pressure in the tank, which is measured by a probe (\textit{Ox Pressure Probe}), but this might also be caused by a problem with the valve between the two tanks (\textit{HeOx Valve Problem}), which is independent of the other variables.  This information is summarised by the QPN in the left panel of Figure \ref{fig:graph3}, which includes positive and negative influences denoted by $S^{+},S^{-}$ respectively, as well as additive synergies denoted by $Y^{+}$.

Of these variables, only the values of the two probes can be measured.  \cite{Druzdzel1993} used their message passing algorithm to investigate the effect of observing a high reading of the \textit{HeOx Temp Probe} on the remaining variables.  The results of the algorithm are displayed in the right panel of Figure \ref{fig:graph3}.  In particular, they conclude that the observation makes larger values of \textit{HeOx Temp}, \textit{High Ox Temp} and \textit{Ox Tank Leak} more likely, while it makes smaller values of \textit{Ox Pressure Probe} more likely.  It also has no effect on \textit{HeOx Valve Problem} due to the independence specified by the graphical structure.

However, we will now provide an example probability distribution which demonstrates that these inferences are incorrect (note that the author has no expertise in the field of this application and so these distributions are for demonstration purposes only).  To construct the probability distribution we begin with the value of \textit{HeOx Temp} which is assumed to follow a uniform distribution between $0$ and $10$.  Then we assume that \textit{HeOx Temp Probe} perfectly measures \textit{HeOx Temp} (i.e. they are equal), but that there is a small probability of there being a fault, in which case the probe records a temperature which is uniform between $5$ and $10$, and is independent of \textit{HeOx Temp}.  Since neither of these variables have other parents in the graph, we can check for positive influences with no other information.  In this distribution, \textit{HeOx Temp} quite clearly positively influences \textit{HeOx Temp Probe} satisfying Definition \ref{def:PosInf} and hence satisfies this assumption of the QPN.  However, \textit{HeOx Temp Probe} does not positively influence \textit{He Ox Temp}. If the probe shows a value of $4$ then \textit{HeOx Temp} is also equal to $4$ with probability $1$. However if the probe shows a value larger than $5$ then there is a small probability that \textit{HeOx Temp} is smaller than $4$.

This demonstrates that, without further assumptions, one cannot conclude, based on an observation of large \textit{HeOx Temp Probe}, that a large value of \textit{HeOx Temp} is more likely in terms of FSD.  Hence the $+$ symbol for \textit{HeOx Temp} in the right of Figure \ref{fig:graph3} must be replaced by a $?$ since the effect is undetermined.

Additionally, all other non-zero inferences in the right of Figure \ref{fig:graph3} are effected by this, even though they do not directly utilise the incorrect symmetry property.  This is because these conclusions rely on the preceding incorrect inference.  Hence all non-zero signs from the message passing algorithm must be replaced by $?$.  This demonstrates the wide extent to which inferences can be effected.

All other current inference algorithms \citep{Wellman1990,Wellman1990a,Renooij1998,Renooij2002,VanKouwen2009} would arrive at the same conclusion as \cite{Druzdzel1993} and therefore contain the same incorrect inferences.

Within this example, one method for rectifying the situation is to redefine all variables as binary variables.  For example, the probe might only indicate whether the temperature is above some critical threshold.  In this way all inferences are maintained.

\section{Discussion and future work}\label{sec:disc}

In this paper we have demonstrated that for non-binary QPNs, many inferences that were previously believed to be true are not mathematically correct.  Without additional work, this severely weakens the conclusions that one is able to make from a QPN.  However, there are a number of ways in which this might be mitigated.

First, it should be stressed that the mistakes highlighted do not effect QPNs with all binary variables.  This is the case when all variables correspond to events and binary QPNs make up a large part of the applied examples.  Additionally, much of the literature of QPNs only focuses on binary variables further highlighting their importance.  As mentioned above, for non-binary QPNs it may be possible to redefine the variables as binary by considering thresholds suitable for the specific application.  This is particularly relevant when using a message passing algorithm because the first step of the algorithm is to determine if an observation is considered large or small.  This implies that there is some threshold to distinguish between large and small values which can be used to define the new binary variables.

For non-binary QPNs, current inference algorithms might still be used by considering a weaker form of inference.  Currently, inference in a QPN is defined via FSD in the same way as positive and negative influences.  For example, the $+$ sign on \textit{HeOx Temp} on the right of Figure \ref{fig:graph3} implies that the conditional distribution given the observation FSD all conditional distributions given smaller observations.  We have demonstrated in this paper that such inferences may be incorrect.  However, there might be weaker forms of inference that can be made.  For example, we can conclude that the two variables are associated (see \cite{Esary1967}, Theorem 4.3).  The strength of conclusions that can be made involve investigating what can be deduced about the set of conditional distributions of $X_1 \mid X_2=x_2$ assuming that $X_1$ positively influences $X_2$ and are a topic for future research.

Alternatively, $S^+(i,j)$ might be redefined using a different definition of positive dependence which does satisfy the symmetry required of the inference algorithms.  However, care must be taken when choosing the positive dependence definition that other important properties of QPN remain valid (for example \cite{Wellman1990}, Theorem 4.3) and they retain the interpretability which is one of the key strengths of QPNs.  For example, the MLRP would be an obvious candidate for a new positive dependence definition.  However, the MLRP is a very strong form of positive dependence and therefore many distributions which would commonly be interpreted as being positive dependent do not satisfy the MLRP \citep{Fallat2017}.  Often the structure of a QPN is elicited from expert judgements and using a less explicit definition of positive dependence might complicate this process.  It could even lead to incorrect inferences because the elicited information might not match the mathematical definition of the QPN.  An alternative approach might be to define $S^+(i,j)$ such that $X_i$ positively influences $X_j$ \textit{and} $X_j$ positively influences $X_i$ so that it becomes a symmetric property.  As discussed in this paper, there are many possible definitions for positive dependence and future research is required to determine which definitions best suit the QPN framework allowing for both strong inferences and simple interpretation.

There has been other research in combining graphical models with positive dependence assumptions, primarily with the multivariate version of the TP$_2$ property (MTP$_2$).  \cite{Fallat2017} gave a comprehensive review of the MTP$_2$ property along with significant research into its combination with conditional independence models.  The MTP$_2$ property was also researched in Gaussian graphical models by \cite{Slawski2015}, \cite{Lauritzen2017} and \cite{wang2020learning}. Recently, work on Gaussian distributions has been extended to another positive dependence definition more akin to positive influences in combination with DAGs \citep{lodhia2023positivity}.  Additional research on positive dependence and graphical models is by \cite{plajner2020learning} who used monotonicity conditions akin to positive and negative influences to improve learning in Bayesian networks.

\section*{Acknowledgements}
The research by JSC was supported in part by the MIUR Excellence Department Project awarded to Dipartimento di Matematica, Università di Genova, CUP D33C23001110001 and the 100021-BIPE 2020 grant.

\newpage
\bibliographystyle{plainnat}
\bibliography{QPNs}

\end{document}